\documentclass[pdflatex,sn-nature]{sn-jnl}
\usepackage{graphicx}%
\usepackage{multirow}%
\usepackage{amsmath,amssymb,amsfonts}%
\usepackage{amsthm}%
\usepackage{mathrsfs}%
\usepackage[title]{appendix}%
\usepackage{xcolor}%
\usepackage{textcomp}%
\usepackage{manyfoot}%
\usepackage{booktabs}%
\usepackage{algorithm}%
\usepackage{algorithmicx}%
\usepackage{algpseudocode}%
\usepackage{listings}%

\usepackage{geometry}
\theoremstyle{thmstyleone}%

\theoremstyle{thmstyletwo}%

\theoremstyle{thmstylethree}%

\begin{document}

\title[Article Title]{Learning Diverse Natural Behaviors for Enhancing the Agility of Quadrupedal Robots}

\author[1]{\fnm{Huiqiao} \sur{Fu}}

\author[1]{\fnm{Haoyu} \sur{Dong}}

\author[1]{\fnm{Wentao} \sur{Xu}}

\author[1]{\fnm{Zhehao} \sur{Zhou}}

\author[1]{\fnm{Guizhou} \sur{Deng}}

\author[1]{\fnm{Kaiqiang} \sur{Tang}}

\author[2]{\fnm{Daoyi} \sur{Dong}}

\author*[1]{\fnm{Chunlin} \sur{Chen}}\email{clchen@nju.edu.cn}

\affil[1]{\orgdiv{Department of Control Science and Intelligent Engineering}, \orgdiv{School of Management and Engineering}, \orgname{Nanjing University}, \orgaddress{\city{Nanjing}, \country{China}}}

\affil[2]{\orgdiv{Australian Artificial Intelligence Institute}, \orgname{University of Technology Sydney}, \orgaddress{\city{Sydney}, \country{Australia}}}


\abstract{Achieving animal-like agility is a longstanding goal in quadrupedal robotics. While recent studies have successfully demonstrated imitation of specific behaviors, enabling robots to replicate a broader range of natural behaviors in real-world environments remains an open challenge. Here we propose an integrated controller comprising a Basic Behavior Controller (BBC) and a Task-Specific Controller (TSC) which can effectively learn diverse natural quadrupedal behaviors in an enhanced simulator and efficiently transfer them to the real world. Specifically, the BBC is trained using a novel semi-supervised generative adversarial imitation learning algorithm to extract diverse behavioral styles from raw motion capture data of real dogs, enabling smooth behavior transitions by adjusting discrete and continuous latent variable inputs. The TSC, trained via privileged learning with depth images as input, coordinates the BBC to efficiently perform various tasks. Additionally, we employ evolutionary adversarial simulator identification to optimize the simulator, aligning it closely with reality. After training, the robot exhibits diverse natural behaviors, successfully completing the quadrupedal agility challenge at an average speed of 1.1 m/s and achieving a peak speed of 3.2 m/s during hurdling. This work represents a substantial step toward animal-like agility in quadrupedal robots, opening avenues for their deployment in increasingly complex real-world environments.}


\keywords{Imitation learning, Locomotion, Natural behaviors, Quadrupedal robots}

\maketitle

\section{Introduction}\label{sec1}

Over hundreds of millions of years, quadrupedal animals have evolved diverse behavioral patterns to cope with survival challenges, allowing them to navigate a wide variety of obstacle-filled environments. To showcase such remarkable agility, global competitions are regularly held, one of the most renowned being the dog agility at the Crufts dog show~\cite{Kennel2024Crufts}, where dogs swiftly maneuver through obstacles guided by handlers. Inspired by these natural abilities, quadrupedal robots have been developed to assist humans in complex and hazardous tasks~\cite{Arm2023Scientific, Miki2022Learning}. However, current quadrupedal robots still fall short of the agility demonstrated by animals, restricting their effectiveness and applicability. In this paper, we propose an integrated controller that significantly enhances the agility of quadrupedal robots, enabling them to successfully complete complex agility tasks.

Recent neuroscience studies reveal that quadrupedal locomotion is governed by both high-level neural regions (cerebral cortex and basal ganglia) for sensory integration and motion planning, and low-level neural regions (brainstem and spinal cord) for posture, balance, and basic gait patterns~\cite{Sten2020Current}. Translating this sophisticated biological control system into robots poses significant challenges. Robots typically have fewer joints, limited actuators, and lack biological structures, resulting in inherent mechanical constraints. Additionally, robots frequently operate in dynamically balanced states, making real-time responsiveness and robustness challenging, especially when replicating diverse natural behaviors. Furthermore, effective obstacle navigation requires integrating high-dimensional sensory inputs for motion planning -- tasks naturally performed by animals but still challenging for robots.

Traditional methods for controlling quadrupedal robots, such as model predictive control and whole-body control~\cite{Di2018Dynamic, Bellicoso2018Dynamic, Fahmi2019Passive, Kim2019Highly}, typically rely on simplified models to optimize foot contacts and joint torques, limiting their performance in highly dynamic and complex tasks. Advanced methods, particularly Deep Reinforcement Learning (DRL) and Imitation Learning (IL), have emerged to overcome these limitations~\cite{sutton2018reinforcement, tang2024deep, hussein2017imitation}. DRL effectively learns robust behaviors through trial and error, handling high-dimensional sensory inputs without explicit models. It has been successfully applied to challenging environments such as stairs~\cite{siekmann2021blind, vogel2024robust, han2024lifelike}, pipes~\cite{jang2022development}, piles~\cite{fu2021deep, wang2023hierarchical}, rough terrain~\cite{Choi2023Learning, fankhauser2018robust, tsounis2020deepgait, cheng2024extreme, rudin2022learning}, and outdoor environments~\cite{Lee2020Learning, Miki2022Learning, jin2022high}. However, designing suitable reward functions in DRL is complex and often leads to unnatural behaviors. In contrast, IL techniques like Behavior Cloning (BC) replicate expert demonstrations without explicit rewards~\cite{torabi2018behavioral, peng2020learning, pearce2023imitating} but often requires extensive data to mitigate compounding errors. In contrast, Adversarial Imitation Learning (AIL) methods~\cite{finn2016connection, fu2018learning, ho2016generative, orsini2021matters} leverage both offline data and online interaction, where compounding errors have been shown to be independent of trajectory length~\cite{xu2022understanding}, thus reducing data requirements. AIL has demonstrated impressive results in physics simulations, including training humanoid characters to perform highly dynamic, natural behaviors such as running, jumping, and backflips~\cite{peng2021amp}. Moreover, integrating AIL with latent variable models has enabled the unsupervised imitation of diverse, multimodal behaviors.~\cite{li2017infogail, peng2022ase, dou2023c}. However, disentangling multimodal behavior features remains challenging. The absence of inductive biases in the model and data makes it difficult to learn desired decodable factors from unstructured data, often leading to mode collapse where the learned policies become difficult to control and exhibit limited behavioral diversity. Furthermore, although numerous studies have demonstrated high dynamic physical behaviors of simulated agents~\cite{peng2021amp, peng2022ase, dou2023c}, transferring these behaviors to the real world remains a major challenge due to the the sim-to-real gap~\cite{zhao2020sim}. Domain randomization (DR) partially addresses this gap by varying simulation parameters~\cite{peng2018sim, tobin2017domain, andrychowicz2020learning}, but achieving effective randomization levels is difficult and typically requires extensive tuning. Recent studies have explored using real-world data to automatically optimize simulator parameters~\cite{dong2024easi, memmel2024asid}, offering new insights into solving the sim-to-real problem.

In this work, inspired by behavioral neuroscience, we introduce an integrated controller to significantly enhance the agility of quadrupedal robots. This controller consists of two main components: the Basic Behavior Controller (BBC) and the Task-Specific Controller (TSC). The BBC, functioning like low-level neural regions, utilizes proprioceptive perception as input and learns diverse behaviors from real dog motion capture data through semi-supervised imitation learning. The TSC, mimicking high-level neural regions, integrates depth images as exteroceptive perception and applies task-driven deep reinforcement learning for various tasks. Compared to existing controllers, we propose three key improvements. First, we develop a semi-supervised variation of Information Maximizing Generative Adversarial Imitation Learning (InfoGAIL)~\cite{li2017infogail, spurr2017guiding, fu2023ess} to train the BBC for controllable multimodal natural behaviors. Due to challenges such as multimodality and data imbalance, extracting behavior features from raw data in an unsupervised manner is challenging. Our semi-supervised framework uses a small amount of labeled data to guide the disentanglement of multimodal behaviors. Specifically, we use the latent skill variable to represent five common dog behavior modes: walk, pace, trot, canter, and jump, and the latent shifting variable to model continuous style variations. By maximizing the variational lower bound of the mutual information between the latent variables and state transitions, a single policy is able to learn diverse controllable behaviors. Additionally, to address the imbalance of behavior modes in the raw motion capture data, the latent skill distribution is dynamically adjusted during training. Combined with the Regularized Information Maximization (RIM) technique~\cite{krause2010discriminative}, the intrinsic information in the unlabeled imbalanced data can be effectively utilized. Second, we employ Evolutionary Adversarial Simulator Identification (EASI) technique~\cite{dong2024easi} to optimize simulator parameters with limited real-world data, thereby reducing the sim-to-real gap. Here, simulator parameter identification is framed as a search problem, solved by an Evolutionary Strategy (ES) acting adversarially against a neural discriminator that scores transitions based on their realism. The entire parameter search process is efficiently parallelized on a GPU-based physics simulator, completing less than 10 minutes. Third, the TSC is trained through a privileged learning architecture~\cite{chen2020learning, Lee2020Learning, cheng2024extreme} to effectively leverage the BBC for complex downstream tasks: a teacher policy first uses privileged information and Hybrid-PPO~\cite{ijcai2019p0316} to maximize task rewards in a hybrid action space; then, a student policy, relying solely on depth images, learns to imitate the teacher’s behavior while incorporating a self-supervised contrastive objective~\cite{grill2020bootstrap} to improve robustness against real-world perception noise.

With only simple depth perception, our integrated controller enabled the robot to autonomously switch between natural behavior modes and successfully complete the quadrupedal agility challenge for the first time, achieving an average speed of 1.1 m/s and a peak speed of 3.2 m/s in the hurdling task. Moreover, the proposed quadrupedal agility challenge provides a benchmark platform that inspires robotic enthusiasts and accelerates progress toward robots with agility rivaling -- or even surpassing -- that of real dogs.

\section{Results}\label{sec2}

\begin{figure}
	\centering
	\includegraphics[width=\textwidth]{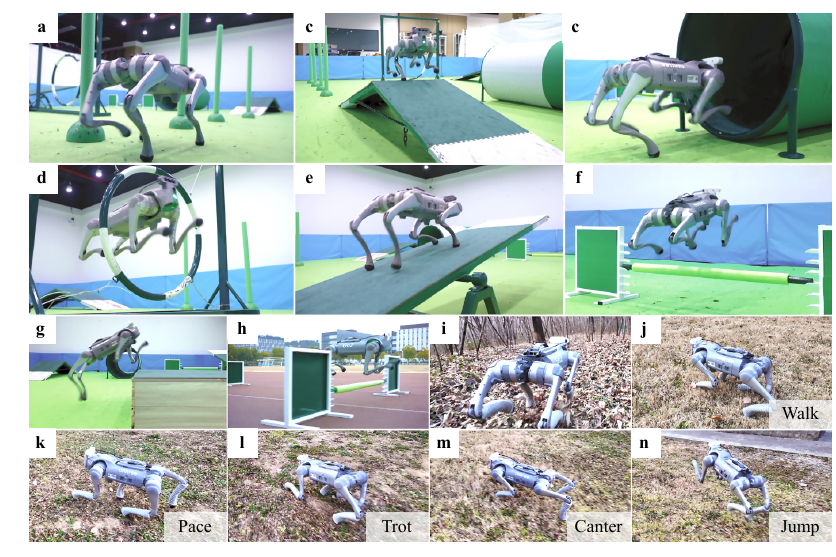}
	\caption{\textbf{Deployment of the proposed controller.} \textbf{a}-\textbf{h}, Environments with various obstacles. \textbf{i}, Wild environment. \textbf{j}-\textbf{n}, Natural behavioral modes.}
	\label{fig:overview}
\end{figure}

Based on our proposed integrated controller, we evaluated the agility of the quadrupedal robot across diverse obstacle environments, as shown in Fig.\ref{fig:overview} and Movie 1. These evaluations included the quadrupedal agility challenge (Fig.\hyperref[fig:overview]{\ref*{fig:overview}a-f}), featuring six randomly placed obstacles -- A-frame, bar-jump, poles, seesaw, tire-jump, and tunnel -- as well as box jumping (Fig.\hyperref[fig:overview]{\ref*{fig:overview}g}), hurdling (Fig.\hyperref[fig:overview]{\ref*{fig:overview}h}), and tests conducted in the wild environment (Fig.\hyperref[fig:overview]{\ref*{fig:overview}i}). Leveraging diverse behavioral modes (Fig.\hyperref[fig:overview]{\ref*{fig:overview}j-n}), the robot efficiently completed each task. All experiments were performed using a Go2 quadrupedal robot equipped solely with an Intel RealSense D435i for depth perception. The robot successfully completed the entire quadrupedal agility challenge with an average speed of 1.1 m/s, reaching a peak speed of 3.2 m/s in the hurdling task. Our results demonstrate that the robot can autonomously navigate complex scenarios using depth perception alone, highlighting significant progress toward robots capable of agile, natural, and versatile performance in real-world applications.

\subsection{Quadrupedal agility challenge}
\begin{figure}
	\centering
	\includegraphics[width=0.95\textwidth]{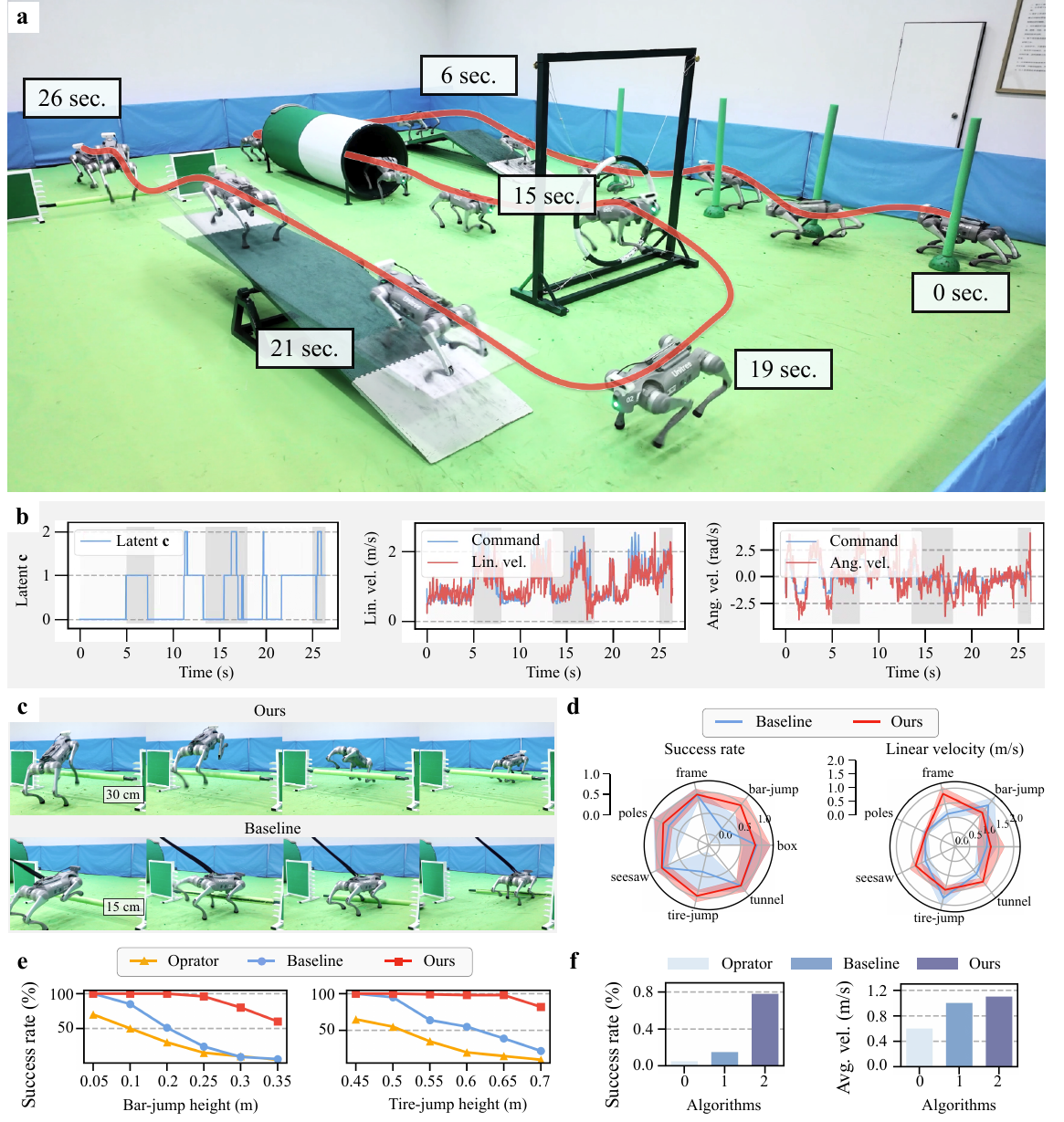}
	\caption{\textbf{Evaluation in the quadrupedal agility challenge.} \textbf{a}, The full motion sequence from start to end. \textbf{b}, Command variations during the motion process, with the gray-and-white shaded areas representing different obstacle stages. \textbf{c}, A comparison of our method and the baseline method~\cite{cheng2024extreme} through the bar-jump. \textbf{d}, A comparison of success rates and average speeds between our method and the baseline method in single-obstacle environments. \textbf{e}, Success rates for different obstacle heights. The ``Operator'' refers to a human remotely controlling the BBC. \textbf{f}, Success rates and average speeds of different methods in the full quadrupedal agility challenge.}
	\label{fig:agility}
\end{figure}

\begin{table}
    \centering
    \caption{\textbf{Specifications of Obstacles and the Robot.}}
    \begin{tabular}{lccc}
        \hline
        Obstacle/Robot & Length (m) & Width (m) / Diameter (m) & Height / Slope \\ 
        \hline
        A-frame      & 2.9  & 0.6  & 13° \\
        Seesaw       & 3.0  & 0.6  & 10° \\
        Poles        & 1.0    & 0.06 (diameter)  & - \\
        Bar-jump     & 1.2  & -    & 0.05 - 0.35 \\
        Tire-jump    & -    & 0.8 (diameter)  & 0.45 - 0.7 \\
        Tunnel       & 2.0  & 0.8 (diameter)  & - \\
        Box       & 1.5  & 1.5  & 0.3 \\
        \hline
        Robot        & 0.7  & 0.31  & 0.34 \\
        \hline
    \end{tabular}
    \label{tab:obstacle_robot_specs}
\end{table}

Inspired by dog agility, we construct the quadrupedal agility challenge in a 7 × 10 m space (Fig.\hyperref[fig:agility]{\ref*{fig:agility}a}), including six types of obstacles: A-frame, bar-jump, poles, seesaw, tire-jump, and tunnel. The order of obstacles, as well as their positions and yaw angles, are randomly configured. The dimensions of each obstacle and the robot can be found in Table~\ref{tab:obstacle_robot_specs}.

The robot's task is to start from the beginning, navigate all obstacles in the shortest time, and reach the end without knocking over any obstacles. Our controller relies solely on simple depth perception, without requiring precise global localization, and remains robust to random environmental changes and disturbances in long-horizon tasks. The robot completes the entire agility challenge in 26 seconds, achieving an average speed of 1.1 m/s. The TSC processes depth images at 50 Hz to generate commands for the BBC, with depth values clipped between 0.3 and 4 meters. Fig.\hyperref[fig:agility]{\ref*{fig:agility}b} left demonstrates the autonomous switching of behaviors by changing latent skill variable $\mathbf{c}$. The alternating white and gray areas represent different obstacle phases. Since the goal of the quadrupedal agility challenge is to achieve faster movement, the TSC uses only three behaviors from the BBC (optional): trot ($\mathbf{c}=0$), canter ($\mathbf{c}=1$), and jump ($\mathbf{c}=2$). Fig.\hyperref[fig:agility]{\ref*{fig:agility}b} middle shows the linear velocity commands and the estimated linear velocity, while Fig.\hyperref[fig:agility]{\ref*{fig:agility}b} right presents the angular velocity commands and the measured angular velocity, highlighting the BBC's rapid adaptation to changing inputs. The robot uses trot at lower speeds and for sharp turns, while canter is favored at higher speeds, and can jump over obstacles when necessary.

We evaluate the proposed controller against the state-of-the-art baseline~\cite{cheng2024extreme}. To better quantify the performance, we first conduct experiments in isolated obstacle environments. Our controller achieves an average success rate close to 100\% across all obstacle types while maintaining a higher speed (Fig.\hyperref[fig:agility]{\ref*{fig:agility}d}). The baseline, lacking natural behaviors, could navigate obstacles that did not require jumping but struggled to clear obstacles like the bar-jump without contact (Fig.\hyperref[fig:agility]{\ref*{fig:agility}c}).  In Fig.\hyperref[fig:agility]{\ref*{fig:agility}d}, we add a box to help the baseline step onto obstacles using foreleg support. Our controller, which utilizes natural behaviors, achieves a higher speed and success rate. Fig.\hyperref[fig:agility]{\ref*{fig:agility}e} compares success rates for different obstacle heights, with the bar-jump ranging from 0.05 to 0.35 m and the tire-jump from 0.45 to 0.7 m. The ``Operator'' refers to a human remotely controlling the BBC. Our controller consistently outperforms baseline methods across different conditions. We further test the controllers in the full quadrupedal agility challenge (Fig.\hyperref[fig:agility]{\ref*{fig:agility}f}), where obstacle placement and order are randomized. The baseline exhibits lower success rates due to the absence of natural behaviors. Human operators, constrained by reaction time, struggle to execute precise command transitions in real time, making it difficult to complete the full task. With our controller, the success rate exceeds 78\%. Notably, due to the  historical depth vision encoding, the robot accurately anticipated the next obstacle’s information, even when it was not visible in the current frame.

\subsection{Natural jump for high-speed hurdling}
\begin{figure}
	\centering
	\includegraphics[width=0.7\textwidth]{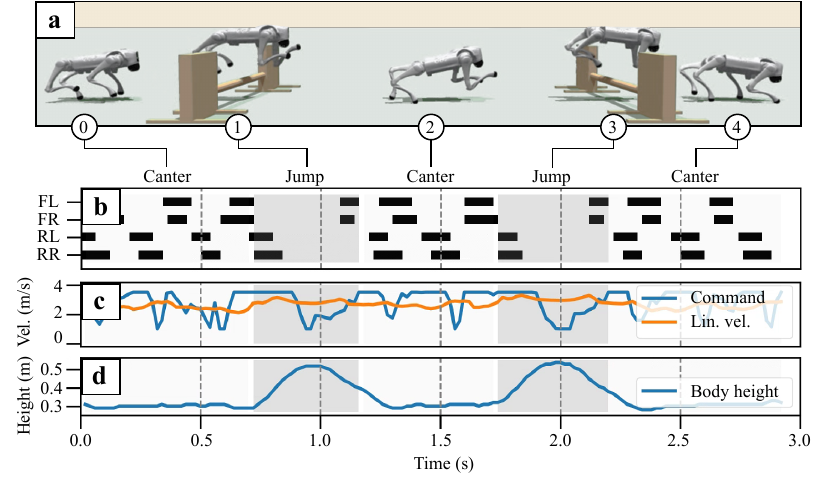}
	\caption{\textbf{Natural jump for high-speed hurdling.} \textbf{a}, Two jumping phases. \textbf{b}, Contact sequence of each leg during hurdling. \textbf{c}, Command and actual linear velocity variations. \textbf{d}, Body height variation.}
	\label{fig:hurdle}
\end{figure}
We evaluate the proposed controller in high-speed motion through a hurdling task. This task consists of four bar-jump obstacles, with spacing randomly set between 2.5 m and 3.5 m, resulting in an average total length of approximately 15 m. Fig.\hyperref[fig:hurdle]{\ref*{fig:hurdle}a} illustrates two jumping phases, while the full process is available in Supplementary Video 3. Fig.\hyperref[fig:hurdle]{\ref*{fig:hurdle}b} shows the contact sequence of each leg during hurdling, where the quadrupedal robot autonomously transitions from canter to jump at the appropriate moment using depth perception. Fig.\hyperref[fig:hurdle]{\ref*{fig:hurdle}c} and Fig.\hyperref[fig:hurdle]{\ref*{fig:hurdle}d} present the velocity and body height variations, respectively. The robot achieves a peak speed of 3.2 m/s, and reaches a maximum body height of 0.54 m during jumps. By adjusting the velocity command and the latent shifting variable, the robot can adapt its stride before and after jumps, which is crucial for maintaining high-speed motion and avoiding obstacle collisions.

\subsection{Evaluating multimodal behaviors of BBC}
\begin{figure}
	\centering
	\includegraphics[width=\textwidth]{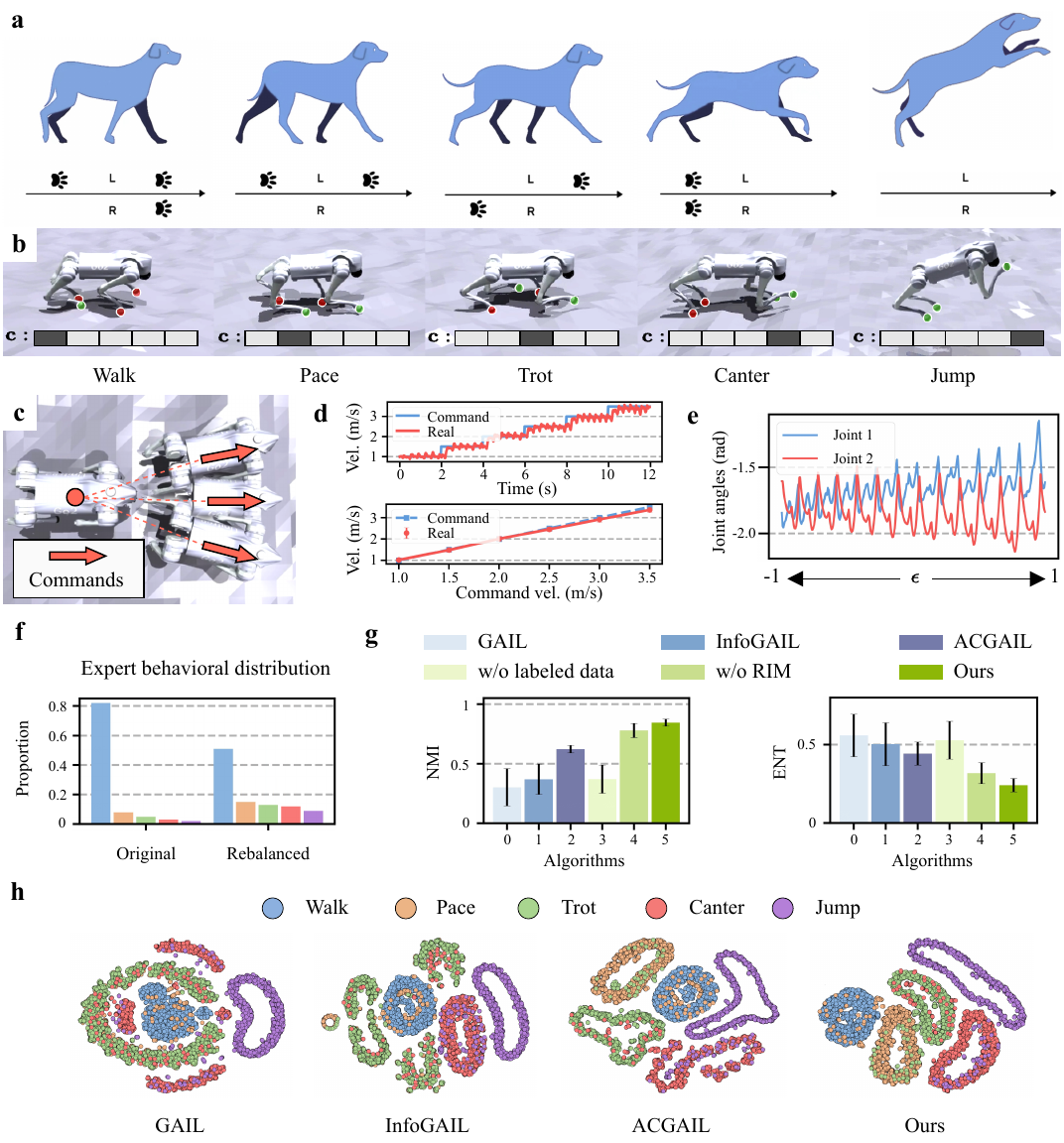}
	\caption{\textbf{Evaluation of the BBC's multimodal behaviors.} \textbf{a}, Common dog behaviors. \textbf{b}, Learned multimodal behaviors for the quadrupedal robot. \textbf{c}, Illustration of command following. \textbf{d}, Command following error. \textbf{e}, The impact of latent shifting variable changes. \textbf{f}, The proportions of the 5 behavior modes in motion capture data. \textbf{g}, NMI ($\uparrow$) and ENT ($\downarrow$) of different algorithms. \textbf{h}, Visualization of behavioral features across different algorithms.}
	\label{fig:bbc}
\end{figure}
Through long-term evolution, dogs have developed a diverse set of basic behaviors to adapt to changing environments. These basic behaviors are controlled independently of the high-level neural regions in their nervous systems and exteroceptive perception. Fig.\hyperref[fig:bbc]{\ref*{fig:bbc}a} illustrates common dog behaviors, including walk, pace, trot, canter, and jump. To replicate such basic behaviors on quadruped robots, we utilize a semi-supervised variation of InfoGAIL to learn a general BBC from a large amount of unlabeled raw motion capture data (with less than 5\% of the data being labeled). By altering the input latent variables of the BBC, the robot can generate a variety of dog-like behaviors. Fig.\hyperref[fig:bbc]{\ref*{fig:bbc}b} shows how the robot can switch between five different behaviors by adjusting the latent skill variable $\mathbf{c}$. In addition to discrete behavior modes, the latent shifting variable $\epsilon$ captures continuous behavior styles in the motion capture data. By adjusting $\epsilon$ continuously between -1 and 1, the robot can produce variations of these behavior styles. Fig.\hyperref[fig:bbc]{\ref*{fig:bbc}e} illustrates how the calf joint angle of the robot’s left front and rear legs changes in response to adjustments in $\epsilon$. To better perform a variety of tasks, the BBC is also trained to respond to speed commands. Fig.\hyperref[fig:bbc]{\ref*{fig:bbc}s} shows the tracking error of the robot's actual linear speed in response to different linear speed command inputs (ranging from 1 to 3.5 m/s) for canter behavior. The robot can quickly respond to speed command inputs, with an average linear speed tracking error of around 0.04 m/s. Before training, we duplicated the labeled data to balance the significant differences in data volume among the behavior categories in the original motion capture data. The proportions of the five behavior modes are illustrated in Fig.\hyperref[fig:bbc]{\ref*{fig:bbc}f}.

We have quantitatively compared our method with several commonly used GAIL variants, including unsupervised GAIL~\cite{ho2016generative} and InfoGAIL~\cite{li2017infogail}, as well as supervised ACGAIL~\cite{lin2018acgail}. In InfoGAIL, we refer to ASE~\cite{peng2022ase} and introduce an additional diversity objective to maximize the diversity of the policy behaviors. Additionally, we conduct two ablation experiments by removing labeled data and RIM to demonstrate the effectiveness of the proposed improvements. To evaluate the diversity of the behaviors learned by different methods, we use two common metrics from clustering algorithms~\cite{xu2003document}: (a) Average Entropy (ENT) assesses two aspects: i) whether state-action pairs generated for a given latent variable belong to the same ground-truth class, and ii) whether each ground-truth class is associated with a unique latent variable. (b) Normalized Mutual Information (NMI) measures the correlation between two clusterings, with values ranging from 0 to 1. A higher NMI value indicates a stronger correlation between the clusterings. The category information in ENT and NMI is output by a pre-trained behavior classifier. The results are shown in Fig.\hyperref[fig:bbc]{\ref*{fig:bbc}g}. The original GAIL struggles to reproduce the complete multimodal behaviors in the dataset due to a lack of inductive bias in data distribution during training, leading to a high ENT and low NMI. InfoGAIL performs better than GAIL by maximizing the mutual information between latent variables and generated trajectories, learning multimodal behaviors unsupervised from the dataset. However, due to the difficulty in controlling the classification features the neural network focuses on, the controller trained with InfoGAIL has poor controllability and is prone to learning undesired behavior categories (e.g., incorrectly treating fast trot and slow trot as two different behavior categories). ACGAIL learns multimodal behaviors from fully labeled data in a supervised manner, but manually annotating large-scale motion capture data is cumbersome. To train ACGAIL, we use the pre-trained classifier to predict labels for the unsupervised data. However, the classifier struggles to distinguish transitional behaviors, and the imbalanced nature of the data limits the final performance of the policy. Our method introduces semi-supervised learning framework, significantly reducing labeling costs compared to ACGAIL while being able to learn discrete categories and capture continuous behavior styles. By incorporating RIM, our method can better handle imbalanced data distributions, achieving the lowest ENT and highest NMI among all baseline methods. To visually compare the learning capabilities of different methods regarding multimodal behaviors, we use t-distributed Stochastic Neighbor Embedding (t-SNE) to reduce the dimensionality of the behavior data generated by these methods, distinguishing different behavior categories by color (Fig.\hyperref[fig:bbc]{\ref*{fig:bbc}h}). The results shows that the different behavior features obtained by our method can be better distinguished.

\subsection{Evaluating TSC in quadrupedal agility challenge}
\begin{figure}
	\centering
	\includegraphics[width=0.8\textwidth]{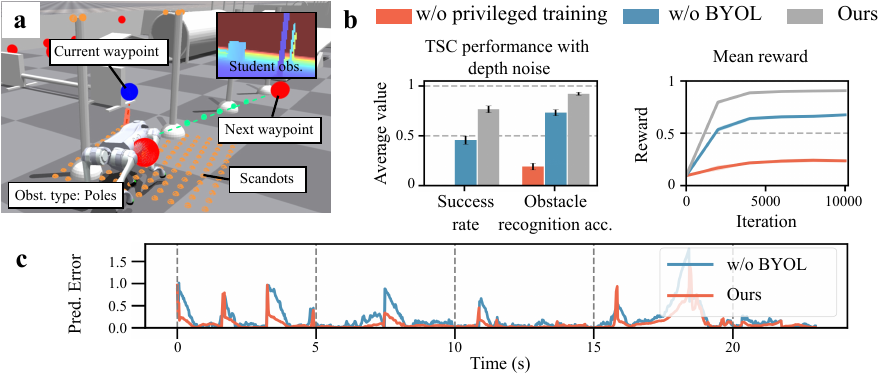}
	\caption{\textbf{Evaluation of the TSC.} \textbf{a}, The privileged information observed by the teacher policy, and the student's observation. \textbf{b}, Comparison between our method and two baseline methods: one without privileged learning and one without BYOL. \textbf{c}, The error between the predicted relative yaw angle and the true value.}
	\label{fig:tsc}
\end{figure}
In the training of TSC, we employ a privileged learning strategy to address sparse rewards in complex tasks. As shown in Fig.\hyperref[fig:tsc]{\ref*{fig:tsc}a}, the teacher policy has access to privileged environmental information, including scandots around the robot, the relative yaw angle between the current and next waypoint, and the type of current obstacles. Since the teacher policy generates both discrete and continuous actions, we train it using the Hybrid-PPO algorithm~\cite{ijcai2019p0316} to maximize task rewards. The student policy, in turn, uses depth images as input to estimate privileged information and mimic the teacher's behavior. It is noteworthy that depth images in real-world environments differ from those in simulation due to noise and interference. To mitigate this, we apply random augmentations to the depth images and utilize a self-supervised learning method called Bootstrap Your Own Latent (BYOL)~\cite{grill2020bootstrap} to enhance the policy's robustness against noise. Fig.\hyperref[fig:tsc]{\ref*{fig:tsc}b} compares our approach with two baselines: one without privileged learning and one without BYOL, in the quadruped agility challenge. The method without privileged learning fails to complete the task, achieving a success rate of 0. Under noise interference, our method outperforms the baseline without BYOL, achieving a higher success rate ($\sim$78\%) and obstacle recognition accuracy ($\sim$92\%), along with the highest task rewards. These results demonstrate that incorporating BYOL improves the policy's robustness to external noise. In addition, we demonstrated the error between the predicted relative yaw angle and the true value under noise interference (Fig.\hyperref[fig:tsc]{\ref*{fig:tsc}c}). Our method shows predictions closer to the true value compared to the baseline without BYOL.

\subsection{Evaluating EASI in sim2real}
\begin{figure}
	\centering
	\includegraphics[width=0.9\textwidth]{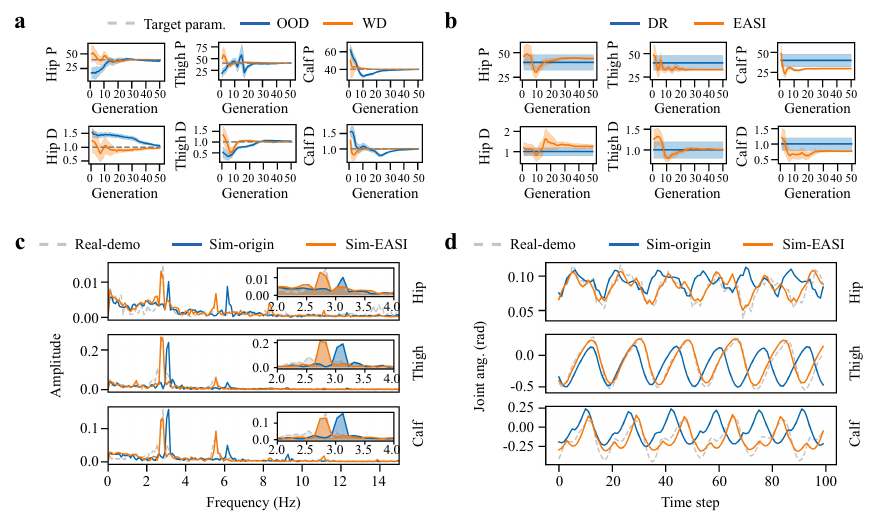}
	\caption{\textbf{Evaluation of the EASI.} \textbf{a}, The parameter search process in the sim-to-sim task. WD means the real parameters (gray dashed line) are within the initial distribution range, while OOD indicates they fall outside this range. \textbf{b},  The parameter search process in the sim-to-real task. \textbf{c}, The joint frequency spectra. \textbf{d}, Joint trajectories.}
	\label{fig:easi}
\end{figure}
Due to the discrepancies between simulation and reality, policies trained in simulation often exhibit suboptimal performance in real-world settings. Conventional DR methods require empirical tuning of parameter randomization ranges, and overly broad or inaccurate ranges can degrade policy performance. In this work, we employ EASI to mitigate this issue by adjusting the parameters of motors at various segments of the quadruped robot within the simulator, thereby reducing the sim-to-real gap. Fig.\hyperref[fig:easi]{\ref*{fig:easi}a} illustrates the parameter search process in a sim-to-sim task, WD indicates that the real parameters (gray dashed line) lie within the initial parameter distribution range, and OOD indicates that the real parameters fall outside this range. After about 50 generations of evolution, all parameters converge close to their true values. Fig.\hyperref[fig:easi]{\ref*{fig:easi}b} depicts the adjustment process of the simulator’s parameter range in a sim-to-real task. To validate the effectiveness of EASI in tuning simulator parameters, we conduct control tasks using the same policy—originally trained via domain randomization—in both the simulator and real world. Fig.\hyperref[fig:easi]{\ref*{fig:easi}c} shows the joint frequency spectra of the robot in the real world and in the simulator before and after adjustment, while Fig.\hyperref[fig:easi]{\ref*{fig:easi}d} presents the corresponding joint kinematic trajectories. With the application of EASI, the simulator becomes more aligned with the real environment, enabling the trained policy to be directly transferred to the real world with minimal performance loss. Notably, the entire parameter search process takes less than 10 minutes and utilizes only 80 seconds of real-world data. The BBC is then fine-tuned in the enhanced simulator for approximately 4000 steps, taking around 2 hours. These improvements ensure that when deployed in the real world, the BBC exhibits agility and robustness more closely resembling its performance in simulation.

\section{Discussion}\label{sec3}

We have proposed an integrated controller in which the BBC acts like low-level neural regions, relying solely on proprioceptive perception to enable a quadrupedal robot to produce diverse, natural behaviors resembling those of real dogs. The TSC functions like high-level neural regions, and uses depth images as exteroceptive perception to generate commands that control the BBC, enabling the robot to efficiently perform a variety of downstream tasks. To bridge the gap between simulation and reality, we used EASI along with a small amount of real-world data to optimize the simulator, aligning it more closely with real-world dynamics. This ensures that the trained policies can be directly deployed, exhibiting agility behaviors in the real world similar to those observed in simulation. With our controller, the quadrupedal robot successfully completed a full quadrupedal agility challenge for the first time, smoothly navigating six randomly placed obstacles at an average speed of 1.1 m/s. During navigation, the robot autonomously adjusted its linear velocity, angular velocity, and behavior modes based on depth images. In the hurdling task, the robot achieved a peak speed of 3.2 m/s with natural canter and jump behaviors. In more complex environments, the controllability of the BBC allows the quadrupedal robot to collaborate with human operators, which can improve task success rates. Our work further enhances the agility of quadrupedal robots, enabling them to overcome various obstacles much like real dogs. Additionally, we have established an engaging quadrupedal agility benchmark, similar to dog agility, and hope it will inspire further research to continuously improve robot agility. Ultimately, we envision these robots competing alongside real dogs and even surpassing them in the future.

\subsection{Possible extensions}\label{subsec2}

In future work, larger-scale real dog outdoor motion capture data can be collected to train a more diverse and robust BBC. Meanwhile, incorporating additional visual information into the motion capture data could help train a more generalizable TSC, improving the robot’s adaptability to real-world environments. Furthermore, the physical parameter distribution in the real world is dynamic; for example, the friction coefficient of contact surfaces varies depending on the material. Using large-scale data, EASI can simulate these changing environmental physical parameters in the simulator, enabling the training of more adaptive policies. Finally, in dog agility challenges, human-dog collaboration plays a crucial role, where handlers are informed of the course in advance and guide the dog through the challenge without physical contact. Currently, we have explored this collaboration mode using remote control; however, more collaboration methods could be introduced, such as voice and gestures. Quadrupedal robots could integrate multimodal information to interpret the operator's intentions, thereby improving their ability to complete more complex tasks.

\section{Methods}\label{sec4}
\subsection{Overview}

\begin{figure}
	\centering
	\includegraphics[width=0.95\textwidth]{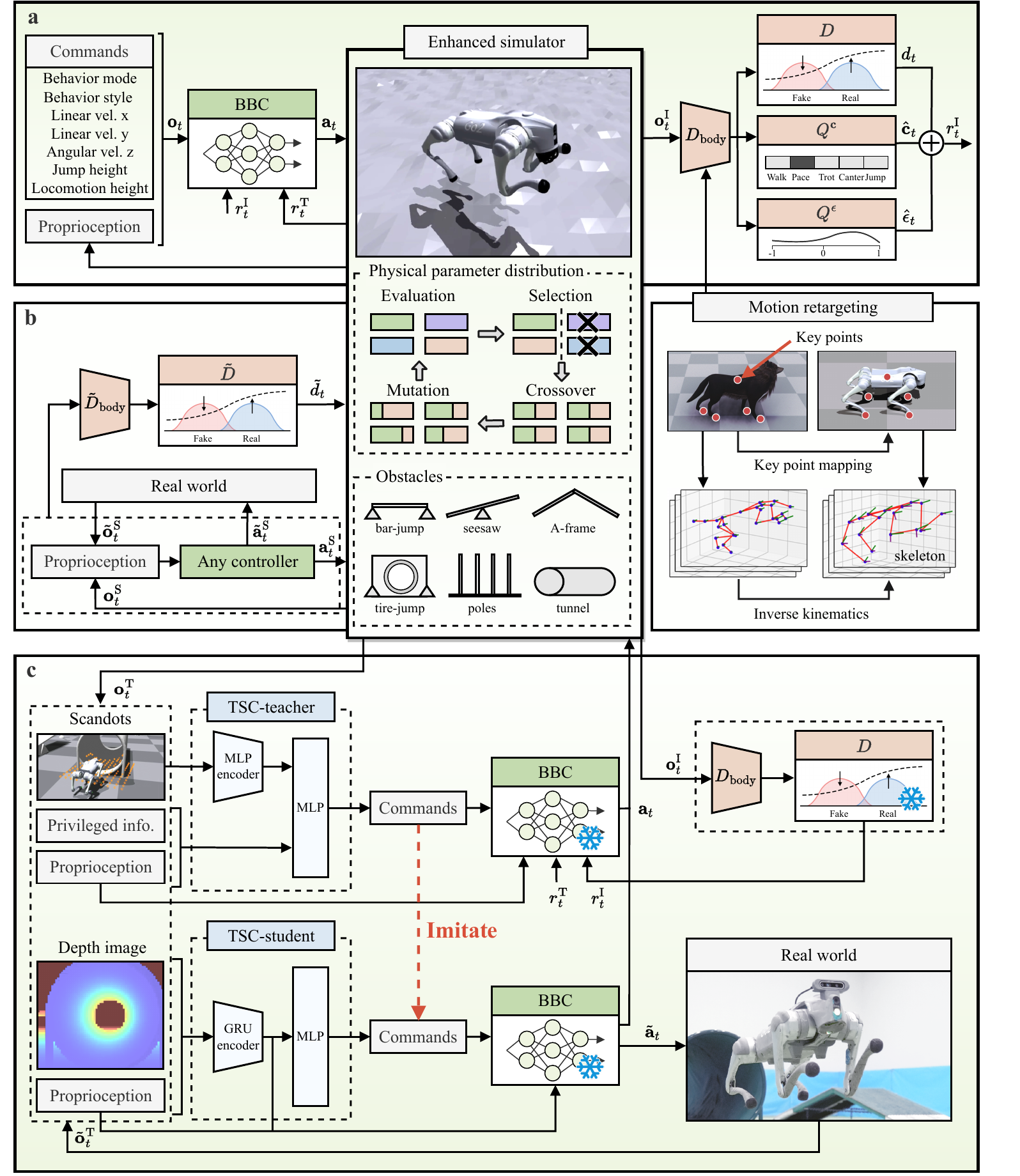}
	\caption{\textbf{Overview of the training process.} \textbf{a}, Basic behavior controller training. \textbf{b}, Simulator identification. \textbf{c}, Task-specific controller training.}
	\label{fig:framework}
\end{figure}

We aim to develop an integrated legged locomotion controller that can be trained in simulation and efficiently transferred to the real world, enabling diverse natural behaviors for quadrupedal agility. We use IssacGym~\cite{makoviychuk2021isaac} as simulator that enables parallel training on GPUs. Our framework comprises three stages, as illustrated in Fig.~\ref{fig:framework}.

In the first stage, a general-purpose BBC is trained using a semi-supervised variant of InfoGAIL to replicate common behaviors observed in real dog motion capture data. The latent skill variable $\mathbf{c}$ is learned in a semi-supervised manner and encodes five discrete behavior modes: walk, pace, trot, canter, and jump. Meanwhile, the latent shifting variable $\epsilon$, learned unsupervisedly, captures continuous behavior styles. The BBC is represented as a neural network to generate behaviors that fool the discriminator while remaining distinguishable by the predictor. This maximizes the variational lower bound of mutual information~\cite{barber2004algorithm} between latent variables and states, thereby enabling a diverse range of multimodal behaviors.

Although the BBC performs well in simulation, its real-world performance is often hindered by the simulation-reality gap. To address this, the second stage uses a small amount of real-world data to enhance the simulator with EASI. Specifically, an arbitrary controller is used to collect motion trajectory data of the quadrupedal robot in the real world. EASI then adjusts the simulator's physical parameters, where the ES acts as a generator in competition with a neural network discriminator. The discriminator evaluates state transitions, assigning higher scores to those closely matching real-world data. These scores serve as the fitness function for the ES, guiding the evolution of physical parameters. Subsequently, the BBC is fine-tuned in the enhanced simulator, which more accurately approximates reality, ensuring efficient real-world transfer with minimal performance loss.

In the final stage, the TSC is trained to generate commands for the BBC to effectively perform downstream tasks. To handle the sparse rewards encountered in complex scenarios, a privileged learning architecture is employed~\cite{chen2020learning}. A teacher policy with access to privileged information, such as scandots, waypoints, and obstacle types, is first trained using hybrid action-space RL. This policy outputs both discrete and continuous commands for the BBC, maximizing both task and imitation rewards. Subsequently, a student policy uses a GRU encoder to process depth images and proprioceptive inputs, predicting the privileged information to mimic the teacher's behavior. The trained student policy, combined with the BBC, forms the integrated legged locomotion controller, which can be directly deployed in the real world to perform a variety of tasks.

\subsection{Problem formulation}
We formulate the control problem of quadrupedal robots as discrete time dynamics, which satisfies a Partially Observable Markov Process (POMDP). At each time step $t$, the policy performs an action $\mathbf{a}_t$, which causes the environment to transition to state $\mathbf{s}_{t+1}$ with probability $\mathcal{P}(\mathbf{s}_{t+1}|\mathbf{s}_{t}, \mathbf{a}_{t})$. At the same time, the agent receives an observation $\mathbf{o}_{t+1}$ with probability $\mathcal{O}(\mathbf{o}_{t+1}|\mathbf{s}_{t+1}, \mathbf{a}_{t})$ and a reward $r_t$. The goal for an RL agent is to find a policy $\pi^{*}$ that maximize the expected reward $\mathbb{E}[\sum_{t=0}^{\infty} \gamma^t r_t]$, where $\gamma \in(0,1)$ is the discount factor.

To imitate behavior from expert demonstrations, GAIL introduces an imitation term $r^{\text{D}}$ to the reward. This term is provided by a discriminator $D(\mathbf{o}^{\text{I}})$, which aims to distinguish state transitions $d^{\pi}(\mathbf{o}^{\text{I}})$ generated by the policy $\pi$ from expert demonstrations $d^{\text{E}}(\mathbf{o}^{\text{I}})$. The discriminator's observation is represented by $\mathbf{o}^{\text{I}}=\{\mathbf{o}^{\text{I}}_{t-H+2}, \cdots, \mathbf{o}^{\text{I}}_{t+1}\}$ with length $H$. To get more stable training and higher quality results, we minimize a discriminator objective similar to Least-Squares Generative Adversarial Networks (LSGAN)~\cite{mao2017least}, as follows:
\begin{equation}
L^{\text{GAIL}}(\pi, D)=\mathbb{E}_{d^{\pi}}\left[(D(\mathbf{o}^{\text{I}})+1)^2\right]+\mathbb{E}_{d^{\text{E}}}\left[(D(\mathbf{o}^{\text{I}})-1)^2\right].
\label{eq_gail_objective}
\end{equation}

The discriminator is trained by solving a least-squares regression problem to predict a score of -1 for samples recorded from the policy and 1 for samples from the demonstration. The imitation reward for the policy is then given by
\begin{equation}
r^{\text{D}} = \max\left[0, 1-0.25(D(\mathbf{o}^{\text{I}})-1)^2\right],
\label{eq_r_gail}
\end{equation}
where the additional offset, scaling, and clipping are applied to bound the reward between $[0, 1]$, as is common practice in previous works~\cite{peng2021amp, tassa2018deepmind}. By iteratively updating $\pi$ and $D$, this least-squares objective ultimately minimizes the Pearson $\chi^2$ divergence between $d^{\pi}$ and $d^{\text{E}}$~\cite{mao2017least}.

\subsection{Basic behavior controller training}
To replicate the behavior of real dogs, we train a general BBC from motion captured data that can generate diverse behaviors by changing the input of latent variables to cope with downstream tasks. We decompose the latent variable into a semi-supervised part $\mathbf{c}$ and an unsupervised part $\epsilon$, where $\mathbf{c}$ is a latent skill variable which is sampled from a categorical distribution, allowing for encoding the same information as the label $y$, and $\epsilon$ a latent shifting variable which is sampled from a continuous uniform distribution, allowing for style shifting within a given skill.

\subsubsection{Semi-supervised skill learning}
We use the latent skill variable $\mathbf{c}$ to represent five behavioral modes: walk, pace, trot, canter, and jump, and seek to maximize the semi-supervised mutual information $I(\mathbf{c}; \mathbf{o}^{\text{I}})$. Since the original mutual information is typically intractable, in practice, it can be approximated using the following variational lower bounds~\cite{spurr2017guiding, fu2023ess}:
\begin{equation}
\begin{aligned}
L_{1}^{\text{SS}}(Q_{1}) &=\mathbb{E}_{p(y), d^{\text{EL}}(\mathbf{o}^{\text{I}}|y)}[\log Q_{1}(y|\mathbf{o}^{\text{I}})]+\mathcal{H}(y) \\ &\leq I(y ; \mathbf{o}^{\text{I}}),
\end{aligned}
\label{eq_l1}
\end{equation}
\begin{equation}
\begin{aligned}
L_{2}^{\text{SS}}(\pi, Q_{2}) &=\mathbb{E}_{p(\mathbf{c}), d^{\pi}(\mathbf{o}^{\text{I}}|\mathbf{c})}[\log Q_{2}(\mathbf{c}|\mathbf{o}^{\text{I}})]+\mathcal{H}(\mathbf{c}) \\ &\leq I(\mathbf{c} ; \mathbf{o}^{\text{I}}),
\end{aligned}
\label{eq_l2}
\end{equation}
where $Q(\cdot|\mathbf{o}^{\text{I}})$ is an approximation predictor of the true posterior $P(\cdot|\mathbf{o}^{\text{I}})$, and the lower bound is tight when $Q=P$. Equation~(\ref{eq_l1}) uses the state transition distribution of the limited labeled expert demonstration $d^{\text{EL}}$ to predict the ground truth label $y$. Equation~(\ref{eq_l2}) utilizes the state transition distribution $d^{\pi}$ produced by a policy $\pi$ to learn the inherent semantic meaning of $y$. Since $p(y)$ and $p(\mathbf{c})$ are independent of $\pi$, $\mathcal{H}(y)$ and $\mathcal{H}(\mathbf{c})$ can be regarded as a constant and does not influence the optimization process. This technique of lower bounding mutual information is known as variational information maximization~\cite{barber2004algorithm}. With $Q_{1} = Q_{2} = Q^{\mathbf{c}}$, we obtain the semi-supervised regularization term:
\begin{equation}
L^{\text{SS}}(\pi, Q^{\mathbf{c}}) = L_{1}^{\text{SS}}(Q^{\mathbf{c}}) + L_{2}^{\text{SS}}(\pi, Q^{\mathbf{c}}).
\label{eq_ss}
\end{equation}

In practice, $L_{1}^{\text{SS}}$ is optimized with respect to $Q^{\mathbf{c}}$ using the cross-entropy loss function, and $L_{2}^{\text{SS}}$ is optimized with respect to $\pi$ using RL. The semi-supervised imitation reward $r^{\text{SS}}$ is then specified by
\begin{equation}
r^{\text{SS}} = \log Q^{\mathbf{c}}(\mathbf{c}|\mathbf{o}^{\text{I}}).
\label{eq_r_ss}
\end{equation}

Unlike previous unsupervised methods~\cite{peng2021amp, peng2022ase, dou2023c}, this objective effectively encourages the policy to generate the desired multimodal behaviors using limited labels. On one hand, it simplifies feature disentanglement and prevents mode collapse; on the other hand, the trained policy exhibits better controllability.

\subsubsection{Unsupervised style learning}
We use the latent shifting variable $\epsilon$ to capture the remaining continuous styles in motion capture data and seek to maximize the unsupervised mutual information $I(\epsilon; \mathbf{o}^{\text{I}})$ by its variational lower bound:
\begin{equation}
\begin{aligned}
L^{\text{US}}(Q^{\epsilon}) &=\mathbb{E}_{p(\epsilon), d^{\pi}(\mathbf{o}^{\text{I}}|\epsilon)}[\log Q^{\epsilon}(\epsilon|\mathbf{o}^{\text{I}})]+\mathcal{H}(\epsilon) \\ &\leq I(\epsilon; \mathbf{o}^{\text{I}}),
\end{aligned}
\label{eq_l3}
\end{equation}
where $\epsilon$ is drawn from a uniform distribution $U(-1, 1)$, $Q^{\epsilon}(\epsilon|\mathbf{o}^{\text{I}})$ is an approximate predictor, and $L^{\text{US}}$ is optimized using the unsupervised imitation reward:
\begin{equation}
r^{\text{US}} = \log Q^{\epsilon}(\epsilon|\mathbf{o}^{\text{I}}).
\label{eq_r_us}
\end{equation}

Note that, since $\epsilon$ is learned in an unsupervised manner, the correspondence between $\epsilon$ and behavior style cannot be determined, performing such subtle operations on $\epsilon$ can be done by the high-level TSC.

\subsubsection{Learning from imbalanced data}
Motion capture data inherently exhibits imbalanced characteristics; for example, dogs spend much more time walking than jumping. Learning multimodal behaviors from such imbalanced data is challenging, and we propose two improvements to alleviate this issue. 

First, most GAN and GAIL algorithms assume that latent variables are sampled from a uniform distribution with fixed parameters. However, this assumption can lead to suboptimal optimization results when dealing with imbalanced data. To align the state transition distribution of the policy with that of the imbalanced expert demonstrations, we use $Q^{\mathbf{c}}$ to predict the empirical label distribution from unlabeled data: $Q^{\mathbf{c}}(\hat{y}) \approx \frac{1}{N}\sum_{i}Q^{\mathbf{c}}(\hat{y} | \mathbf{o}^{\text{I}}_{i})$,  and adopt this distribution as the sampling distribution for latent skill variable $\mathbf{c}$.

Second, to leverage the intrinsic information within the imbalanced unlabeled demonstrations and improve the efficiency of the semi-supervised learning process, we introduce the RIM loss $L^{\text{RIM}}$, a discriminative clustering technique that can automatically identify boundaries or distinctions between categories in unlabeled data~\cite{krause2010discriminative, ghosal2020short}:
\begin{equation}
-\frac{1}{N} \sum_{i} \mathcal{H}\left(Q^{\mathbf{c}}(\hat{y} | \mathbf{o}^{\text{I}}_{i})\right)\vphantom{\frac{1}{N}\sum_{i}}-D_{\text{KL}}\left(Q^{\mathbf{c}}(\hat{y}) \| p(\mathbf{c})\right)\vphantom{\frac{1}{N}\sum_{i}}-R(\psi),
\end{equation}
where the first term is the cluster assumption, corresponding to the idea that datapoints should be classified with large margin~\cite{grandvalet2004semi}. The second term is performed to avoid degenerate solutions. Since we have already set $p(\mathbf{c})=Q^{\mathbf{c}}(\hat{y})$, this term is ultimately eliminated. The last term is a parameter regularization (e.g., L2 regularization) performed to avoid complex solutions.

\subsubsection{Discriminator and predictor objective}
With the above improvements, the objective functions for the discriminator and the predictors $Q^{\mathbf{c}}$ and $Q^{\epsilon}$ eventually become:
\begin{equation}
\min_{D, Q^{\mathbf{c}}, Q^{\epsilon}} L^{\text{GAIL}} - L^{\text{SS}} - L^{\text{US}} - L^{\text{RIM}}.
\label{eq_ss_infogail}
\end{equation}

We demonstrate in the experiment that through this integrated objective, the robot can learn diverse behaviors that include both discrete behavior modes and continuous behavior styles from imbalanced motion capture data, which is crucial for the quadrupedal agility.

\subsubsection{Observation and action}
We define the BBC's observation as $\mathbf{o}$ and the discriminator and predictor's observation as $\mathbf{o}^{\text{I}}$. Specifically, $\mathbf{o}$ consists of:
\begin{itemize}
    \item Command inputs: \{behavior mode, local $x$-axis and $y$-axis linear velocities, local $z$-axis angular velocity, jump height, locomotion height\}.  
    \item Measured proprioception: \{roll and pitch angles, local $z$-axis angular velocity, joint rotations, joint velocities, foot contact states\}.  
    \item Estimated proprioception: \{root height from the ground, local $x$-axis and $y$-axis linear velocities\}.  
    \item Latent physical parameters: \{body mass, friction coefficients, joint PD parameters\}.  
\end{itemize}
Since estimated proprioception cannot be directly measured in the real world, we use a Multi-Layer Perceptron (MLP) to infer it from measured proprioception. Additionally, latent physical parameters are inferred from proprioception history using two 1D convolutional layers.

As a key factor in imitation, $\mathbf{o}^{\text{I}}$ includes the ground-truth proprioception. Additionally, local foot positions are incorporated as supplementary imitation information. The dog motion capture dataset is obtained from \cite{zhang2018mode}. To map the original skeletal animation to a quadrupedal robot, we use the motion retargeting technique. Specifically, five key points, including the center of mass and four feet, are defined on the original skeleton and mapped onto the robot's skeleton, and the joint rotations are computed using inverse kinematics. We manually segment a limited portion of the original motion capture data as labeled data (less than 5\%), with the remainder used as unlabeled data. The action of BBC is defined as the target joint rotations, which are used as inputs for the PD controller.

\subsubsection{Rewards}
The total reward includes task related reward $r^{\text{T}}$ and imitation related reward $r^{\text{I}}$:
\begin{equation}
r = w^{\text{I}}r^{\text{I}} + w^{\text{T}}r^{\text{T}},
\label{eq_reward}
\end{equation}
where $w^{\text{I}}$ and $w^{\text{T}}$ are two coefficients that control the weight of $r^{\text{I}}$ and $r^{\text{T}}$. $r^{\text{I}}$ guides the policy to imitate diverse expert behaviors, and is calculated by~Equations~(\ref{eq_r_gail}), (\ref{eq_r_ss}), (\ref{eq_r_us}) as follows:
\begin{equation}
r^{\text{I}} = r^{\text{D}} + r^{\text{SS}} + r^{\text{US}}.
\end{equation}

The reward $r^{\text{T}}$, which is unrelated to imitation, guides the policy to follow specific command inputs. It includes the following components:

\begin{itemize}
    \item Linear velocity tracking reward: $ r^{\text{T}}_{\text{lin\_vel}} = 2 \cdot \exp(-||\mathbf{v}^{\text{xy}}_{\text{cmd}} - \mathbf{v}^{\text{xy}}||^{2}) $, where $ \mathbf{v}^{\text{xy}}_{\text{cmd}} $ and $ \mathbf{v}^{\text{xy}} $ represent the commanded and actual linear velocities along the $x$ and $y$ axes, respectively.
    \item Angular velocity tracking reward: $ r^{\text{T}}_{\text{ang\_vel}} = 1.5 \cdot \exp(-(\omega^{\text{z}}_{\text{cmd}} - \omega^{\text{z}})^{2}) $, where $ \omega^{\text{z}}_{\text{cmd}} $ and $ \omega^{\text{z}} $ represent the commanded and actual angular velocities along the $z$-axis, respectively.
    \item Jump height reward: $ r^{\text{T}}_{\text{jump}} = 2 \cdot \mathbb{I}_{\text{jump}} \cdot \mathbb{I}_{||h_{\text{cmd}} - h|| < 0.05} $, where $ h_{\text{cmd}} $ and $h$ represent the commanded and actual root heights, respectively. $ \mathbb{I} $ is the indicator function.
    \item Locomotion height reward: $ r^{\text{T}}_{\text{loco}} = 0.1 \cdot \mathbb{I}_{\text{loco}} \cdot \exp(-(h_{\text{cmd}} - h)^{2}) $.
\end{itemize}
We also use the additional regularization terms to achieve smoother and more stable control like~\cite{rudin2022learning}, which are detailed in the Supplementary Section 3. The total reward is maximized by Proximal Policy Optimization (PPO)~\cite{schulman2017proximal}.

\subsection{Evolutionary adversarial simulator identification}
We use EASI to optimize the simulator’s parameter distribution using limited real-world data, effectively narrowing the sim-to-real gap and enabling simulation-trained policies to transfer directly to the real world.

Our goal is to identify the optimal simulator parameter distribution $\Xi^{*}$ that minimizes the discrepancy between the real-world state transition $\mathbb{E}_{\tilde{\xi} \sim \tilde{\Xi}}[P(\tilde{\xi})]$ and the simulated state transition $\mathbb{E}_{\xi \sim \Xi}[P(\xi)]$. We use $\tilde{(\cdot)}$ to represent quantities associated with the real world. In EASI, a discriminator $\tilde{D}$ attempts to distinguish whether state-action transitions originate from the real world or the simulator, assigning higher scores to real-world samples as the fitness function for the ES. As a generator, the ES evaluates, selects, crosses over, and mutates parameter distributions to maximize these scores, as defined by the following objective:
\begin{equation}
\max_{\Xi}\min_{\tilde{D}}\mathbb{E}_{d^{\pi}}\left[(\tilde{D}(\mathbf{o}^{\text{S}}_{t}, \mathbf{a}^{\text{S}}_t,  \mathbf{o}^{\text{S}}_{t+1})+1)^2\right]+\mathbb{E}_{\tilde{d}^{E}}\left[(\tilde{D}(\mathbf{o}^{\text{S}}_{t}, \mathbf{a}^{\text{S}}_t, \mathbf{o}^{\text{S}}_{t+1})-1)^2\right],
\label{eq_easi}
\end{equation}
where $\mathbf{o}^{\text{S}}$ represents the observations of the discriminator $\tilde{D}$, which correspond to the robot's proprioception. $\mathbf{a}^{\text{S}}$ denotes the actions, which can be generated by any controller. In this work, we use the BBC trained with roughly defined DR parameter ranges as the controller to collect state-action trajectory samples in both simulation and the real world. The BBC is then fine-tuned in the enhanced simulator using the optimal parameter distribution $\Xi^{*}$.

\subsection{Task-specific controller training}
To tackle the quadrupedal agility challenge, we train the TSC using a privileged learning architecture. The training process consists of two stages: TSC-teacher policy training and TSC-student policy training.

\subsubsection{TSC-teacher policy training}
We denote the teacher's observation as $\mathbf{o}^{\text{T}}$. The teacher policy $\pi^{\text{TSC}}$ takes privileged information as input, including the following:
\begin{itemize}
    \item Scandots: elevation map of surrounding terrain.
    \item Privileged information: obstacle type, current $\Delta_{\text{yaw}}$ and next $\Delta^{\prime}_{\text{yaw}}$.
    \item Proprioception: measured and estimated proprioception as BBC.
\end{itemize}
Here the obstacle type is a one-hot vector that represents six types of randomly placed obstacles in the quadrupedal agility challenge. $\Delta_{\text{yaw}}$ and $\Delta^{\prime}_{\text{yaw}}$ represent the relative yaw angle between the robot and the next two waypoints.

The output action of the teacher policy includes a mixture of discrete and continuous commands: $\mathbf{a}^{\text{T}}=\{\mathbf{a}^{\text{D}}, \mathbf{a}^{\text{C}}\}$. The discrete part represents the behavior mode, while the continuous part represents the remaining behavior style, speed, and root height commands. To realize hybrid action space policy training, we use Hybrid-PPO~\cite{ijcai2019p0316} to maximize the following rewards:
\begin{equation}
r^{\text{TSC}} = w^{\text{I}}r^{\text{I}} + w^{\text{TSC}}r^{\text{TSC}},
\label{eq_reward_TSC}
\end{equation}
where $r^{\text{I}}$ is computed by the discriminator trained during the BBC's training phase, with the discriminator's parameters fixed. The task-related reward $r^{\text{TSC}}$ includes the following:
\begin{itemize}
    \item Linear velocity tracking reward: $ r^{\text{TSC}}_{\text{lin\_vel}} = 0.3 \cdot \min (\langle\mathbf{v}, \mathbf{d}_{\text{wpt}}\rangle, v_{\text {target }}) $, where $ \mathbf{v} $ is the linear velocity. $\mathbf{d}_{\text{wpt}}=(\mathbf{p}_{\text{wpt}}-\mathbf{p})/(||\mathbf{p}_{\text{wpt}}-\mathbf{p}||)$ is the target waypoint direction, calculated using the current position of the robot $\mathbf{p}$ and the target waypoint position $\mathbf{p}_{\text{wpt}}$.
    \item Yaw tracking reward: $ r^{\text{TSC}}_{\text{yaw}} = 2 \cdot \exp(-||\theta^{\text{z}}_{\text{wpt}} - \theta^{\text{z}}||^{2}) $, where $ \theta^{\text{z}}_{\text{wpt}} $ and $ \theta^{\text{z}} $ denote the yaw angle of the target waypoint and the robot, respectively.
    \item Waypoint reaching reward: $ r^{\text{TSC}}_{\text{wpt}} = 5 \cdot \mathbb{I}_{||\mathbf{p}_{\text{wpt}}-\mathbf{p}|| < 0.4} $ rewards each time the target waypoint is reached.
    \item Termination reward: $r^{\text{TSC}}_{\text{term}} = -50 \cdot\mathbb{I}_{\text{term}}$ penalizes episodes that terminate due to base or hip collisions with obstacles or reaching the maximum step limit.
\end{itemize}
Additional regularization rewards are detailed in the Supplementary Section 4. The above reward terms guide the robot to navigate through randomly placed obstacles while following the designated waypoints.

\subsubsection{TSC-student policy training}
In the student policy, we use depth images as exteroceptive perception combined with proprioceptive perception as inputs to a Gate Recurrent Unit (GRU)~\cite{chung2014empirical}. 
The GRU outputs the predicted privileged information and an embedding of the environment. These are then processed by an MLP to generate the final hybrid student action $\hat{\mathbf{a}}^{\text{T}}=\{\hat{\mathbf{a}}^{\text{D}}, \hat{\mathbf{a}}^{\text{C}}\}$, replicating the behavior of the teacher policy by minimizing the following objective:
\begin{equation}
\mathbb{E}_{\hat{\pi}^{\text{TSC}}}\left[-\sum_{k=1}^K \mathbf{a}^{\text{D}}_k \log \hat{\mathbf{a}}^{\text{D}}_k\right] + \mathbb{E}_{\hat{\pi}^{\text{TSC}}}\left[(\mathbf{a}^{\text{C}}-\hat{\mathbf{a}}^{\text{C}})^{2}\right]
\label{eq_tsc_student}
\end{equation}
where $\hat{\pi}^{\text{TSC}}$ represents the TSC-student policy, and $K$ is the number of discrete actions. The first term corresponds to the cross-entropy loss for discrete actions, while the second term captures the mean squared error for continuous actions.

Depth images in real-world environments often contain noise from various sources, making them different from those in simulation. To address this, we use a self-supervised learning method called Bootstrap Your Own Latent (BYOL)~\cite{grill2020bootstrap}. BYOL learns task-relevant features by maximizing the similarity between two augmented views of the same depth image. We integrate this self-supervised loss into the depth image encoder network and apply probabilistic augmentations such as white noise, background noise, random cropping, edge noise, and Gaussian blur to the depth images. This enhances the encoder's robustness to the complexities of real-world environments.

\bibliography{sn-article}%

\newpage

\renewcommand{\thefigure}{S\arabic{figure}}
\renewcommand{\thetable}{S\arabic{table}}
\renewcommand{\theequation}{S\arabic{equation}}
\renewcommand{\thepage}{S\arabic{page}}
\renewcommand{\thealgorithm}{S\arabic{algorithm}}
\setcounter{figure}{0}
\setcounter{table}{0}
\setcounter{equation}{0}
\setcounter{page}{1}

\begin{center}
\section*{Supplementary Materials for Learning Diverse Natural Behaviors for Enhancing the Agility of Quadrupedal Robots}
\end{center}

\subsection*{S1. Nomenclature}\label{secS1}
\begin{tabular}{cl} 
$\mathbf{s}$ & state \\
$\mathbf{o}$ & observation \\
$\mathbf{a}$ & action \\
$\mathbf{a}^{\text{D}}$ & discrete action \\
$\mathbf{a}^{\text{C}}$ & continuous action \\
$r$ & reward \\
$\pi$ & policy \\
H & observation horizon \\
$D$ & discriminator \\
$Q$ & predictor \\
$\mathbf{c}$ & latent skill variable \\
$\epsilon$ & latent shifting variable \\
$y$ & behavior label \\
$\mathcal{H}$ & entropy\\
$I(\cdot)$ & mutual information \\
$(\cdot)^{\text{I}}$ & imitation quantity \\
$(\cdot)^{\text{T}}$ & task quantity \\
$(\cdot)^{\text{S}}$ & simulator quantity \\
$\tilde{(\cdot)}$ & real-world quantity \\
$\hat{(\cdot)}$ & predicted quantity \\
$(\cdot)^{\text{SS}}$ & semi-supervised quantity \\
$(\cdot)^{\text{US}}$ & unsupervised quantity \\
$\mathbb{I}$ & indicator function\\
$||\cdot||$ & $l_2$ norm\\
\end{tabular}

\subsection*{S2. Training details}\label{secS2}

The training of the integrated controller is divided into three stages: BBC pretraining, BBC fine-tuning, and TSC training. All training processes are conducted in the IsaacGym physics simulation environment.

In the first stage, the simulator's physical parameters are set empirically. We use semi-supervised InfoGAIL to train a multimodal BBC based on motion capture data of common dog behaviors, with less than 5\% of the data labeled. Aside from motion retargeting, the data requires no additional preprocessing, such as cropping, labeling, or alignment. To address category imbalance, labeled data is simply repeated to rebalance minority categories. The training pseudocode is presented in Algorithm~\ref{alg:ss_infogail}.  

In each training iteration $i$, the latent skill variable $\mathbf{c}$ is sampled from a categorical distribution $p_{i}(\mathbf{c})$, and the latent shifting variable $\epsilon$ is drawn from a uniform distribution $p(\epsilon)$. The policy interacts with the environment based on observations and latent variables, generating a trajectory distribution $d^{\pi}$. The discriminator $D$ and predictor $Q$ are updated via gradient descent on Equation~(\ref{eq_ss_infogail}). The policy $\pi$ is then optimized using the PPO algorithm~\cite{schulman2017proximal} to maximize the reward in Equation~(\ref{eq_reward}), allowing it to replicate the multimodal behaviors in the motion capture data. The value function $V$ is updated using $TD(\lambda)$. 

To align with the imbalanced motion capture data, $p(\mathbf{c})$ is estimated by $Q^{\mathbf{c}}$ and updated using the exponential moving average:
\begin{equation}
    p_{i+1}(\mathbf{c}) = \beta p_{i}(\mathbf{c}) + (1-\beta) \frac{1}{N}\sum_{i}Q^{\mathbf{c}}(\hat{y} | \mathbf{o}^{\text{I}}_{i}),
\end{equation}
where $\beta$ is the weight coefficient. The hyperparameters for semi-supervised InfoGAIL are shown in Table~\ref{tab:hype_ss_infogail}. 

In addition to imitating motion capture data, the BBC is also trained to follow task-related command inputs to enhance its controllability. The command ranges for each behavior mode are shown in Table~\ref{tab:command_range}. During training, task commands and latent variables are sampled every six seconds. Additionally, randomly generated rough terrains and random disturbances applied to the robot are introduced to improve the robustness of the BBC.

In the second stage, we use EASI to learn the simulator's parameter distribution from a small set of real-world samples, allowing the simulator to better approximate the real world. The BBC is then fine-tuned in this enhanced simulator. The real-world samples, collected using the pre-trained BBC (or any other controller), consist of 4,000 steps at a control frequency of 50 Hz, providing about 80 seconds of data. The simulator parameter optimization pseudocode is presented in Algorithm~\ref{alg:easi}. 

In each generation, $N$ parameter sets are sampled from the distribution $\Xi^{(i)}$, including the PD parameters for the hip, thigh, and calf joints of the quadrupedal robot, while the known component masses remain fixed. The policy $\pi$ then collects trajectories $d^{\pi}$ in $N$ parallel simulations with the sampled parameters. A batch of state transitions is subsequently used to update the discriminator $\tilde{D}_{k}$ to $\tilde{D}_{k+1}$ following Equation~(\ref{eq_easi}). Finally, the updated discriminator calculates the rewards $r^{(i)}$, which are used by ES to generate the next parameter distribution $\Xi^{(i+1)}$. The hyperparameters for EASI are shown in Table~\ref{tab:hype_easi}. 

In the last stage, the TSC is trained using privileged learning to perform specific tasks. The pseudocode is presented in Algorithm~\ref{alg:priv}. To address the issue of sparse rewards, the teacher policy $\pi^{\text{TSC}}$ is first trained to maximize task reward and imitation reward, with the latter calculated by the discriminator from the BBC. Hybrid-PPO~\cite{ijcai2019p0316} is employed to handle the hybrid action space. Subsequently, the student policy $\hat{\pi}^{\text{TSC}}$ takes depth images as input and outputs actions to mimic the teacher's behavior. The positions of obstacles and the robot are randomly initialized at the beginning of each episode. Additionally, to enhance the robustness of the student policy in real world, we incorporate the BYOL loss~\cite{grill2020bootstrap}. We apply random augmentations to the input depth images, such as rectangular noise, white noise, and Gaussian blur. The hyperparameters for privileged learning are shown in Table~\ref{tab:hype_priv}. 

The networks are implemented using Pytorch~\cite{paszke2019pytorch}, and are trained on a computer with an i5-12600KF CPU and a Nvidia GTX 4080 GPU. We use the parallelized physics simulator IsaacGym~\cite{makoviychuk2021isaac} for training. The BBC is trained using real dog motion capture data from~\cite{zhang2018mode}, which includes approximately 20 minutes of motion data covering common dog behaviors. Training the BBC takes about 4 days on a single GPU, while the TSC-teacher and TSC-student require approximately 10 hours and 14 hours, respectively.

\subsection*{S3. Deployment details}\label{secS3}

We deploy our integrated controller on the Unitree Go2 robot, which is equipped with a Jetson Orin NX, an Intel RealSense D435i depth camera~\cite{IntelRealSense}  and features four limbs comprising 12 joints.

In our real-world experiments, we deploy our integrated controller that incorporates both BBC and TSC on the Jetson Orin NX embedded in the Unitree Go2 robot, achieving stable control frequency at 50 Hz while performing deep image processing.

To ensure that image processing and policy inference are completed within each 20ms cycle, all computations are accelerated using CUDA. Intercommunication among the robot’s various sensors and low-level motors is facilitated via the Cyclone DDS and LCM platforms, with the motors executing torque at a rate of 1000 Hz based on Unitree Go2 integrated PD controllers. Depth images are acquired at a sensor resolution of 848×480, and then these images are subsequently downsampled and cropped to yield a final resolution of 87×58. After normalization in which  each pixel value is scaled to the range from -0.5 to 0.5, the processed depth image serves as the input to the visual encoder network, which operates concurrently with the motion policy at a frequency of 50 Hz. Moreover, owing to the inherent imaging principles of the RealSense D435i, the raw depth images often contain considerable invalid regions and noise. To address these limitations, RGB-Depth Processing module in OpenCV is employed to process the depth map data, effectively enhancing the image quality.

\subsection*{S4. Additional Rewards for BBC}\label{secS4}

To achieve a smoother and more stable controller, we introduce the following regularization rewards for the BBC:
\begin{itemize}
    \item Joint torque reward: $ r^{\text{T}}_{\text{torque}} = -10^{-5} \cdot \sum||\mathbf{\tau}||^{2} $ penalizes excessive joint torques.
    \item Torque change reward: $ r^{\text{T}}_{\text{delta\_torque}} = -10^{-7} \cdot \sum||\Delta_{\mathbf{\tau}}||^{2} $ penalizes rapid changes in joint torque, where $ \Delta_{\mathbf{\tau}} $ is the difference between consecutive steps.  
    \item Joint acceleration reward: $ r^{\text{T}}_{\text{dof\_acc}} = -2.5\times10^{-7} \cdot \sum||\ddot{\mathbf{q}}||^{2} $ penalizes excessive joint acceleration.
    \item Joint position limitation reward: $r^{\text{T}}_{\text{joint\_pos\_limit}} = -0.1 \cdot \sum \left( \max(0, \mathbf{q}_{\text{min}} - \mathbf{q}) + \max(0, \mathbf{q} - \mathbf{q}_{\text{max}}) \right)$ penalizes joint positions that exceed their defined limits.
    \item Joint velocity limitation reward: $  r^{\text{T}}_{\text{joint\_vel\_limit}} = -0.1 \cdot \sum \left( \min(1, \max(0, |\dot{\mathbf{q}}| - \mathbf{q}_{\text{vel\_limit}})) \right)$ penalizes joint velocities exceeding the velocity limit.
    \item Joint torque limitation reward: $  r^{\text{T}}_{\text{joint\_torque\_limit}} = -0.03 \cdot \sum \left( \max(0, |\mathbf{\tau}| - \mathbf{\tau}_{\text{limit}}) \right)$ penalizes joint torques that approach or exceed the torque limit.
    \item Collision reward: $ r^{\text{T}}_{\text{collision}} = -10 \cdot \mathbb{I}_{\text{collision}} $ penalizes collisions involving the thigh and calf.  
    \item Action smoothness reward: $ r^{\text{T}}_{\text{delta\_action}} = -0.1 \cdot ||\Delta_a||^2 $ penalizes abrupt changes in action commands.  
\end{itemize}

\subsection*{S5. Additional rewards for TSC}\label{secS5}

To achieve a smoother and safer TSC, we introduce the following regularization rewards:
\begin{itemize}
    \item Action smoothness reward: $ r^{\text{TSC}}_{\text{delta\_action}} = -0.1 \cdot ||\Delta_{a^{\text{T}}}||^2 $ penalizes abrupt changes in action commands.
    \item Collision reward: $ r^{\text{TSC}}_{\text{collision}} = -20 \cdot \mathbb{I}_{\text{collision}} $ penalizes collisions involving the thigh and calf.
    \item Edge contact reward: $ r^{\text{TSC}}_{\text{feet\_edge}} = - \sum_{i=0}^{4}c^{\text{foot}}_{i}M(p_i) $ penalizes each foot contact with the edge of the obstacles. $c^{\text{foot}_{i}}=1$ if foot $i$ is in contact. $M$ is a boolean function which is 1 if the foot position $p_i$ within the obstacle edge.
\end{itemize}

\subsection*{S6. Neural network architectures}\label{secS6}

The BBC's policy network outputs target positions for the quadrupedal robot's 12 joints. This network is an MLP with 3 hidden layers \{512, 256, 128\}. The input includes commands, measured and estimated proprioception, and latent physical parameters. Estimated proprioception is generated by a two-layer MLP (sizes \{128, 64\}), based on measured proprioception. Latent physical parameters are inferred from proprioception history using two 1D convolutional layers with kernel sizes \{4, 2\} and strides \{2, 1\}.

The TSC-teacher’s policy network outputs both discrete and continuous commands. Discrete commands are probabilities for 5 behaviors, while continuous commands consist of $5 \times 6$ values corresponding to the behaviors. The TSC policy network is an MLP with 3 hidden layers \{512, 256, 128\}. Its input consists of scandots, privileged information, and proprioception. Scandots are an $11 \times 12$ matrix, which is flattened and input into a three-layer MLP encoder (sizes \{128, 64, 32\}), with the 32-dimensional output used as input to the policy network. Estimated proprioception is predicted in the same way as in the BBC.

The TSC-student’s policy network is an MLP with 3 hidden layers \{512, 256, 128\}. It's outputs are identical to the TSC-teacher. Instead of scandots, it uses the current depth map, sized $87 \times 58$ as exteroceptive perception. This depth map is processed with 2 convolutional layers (kernel sizes \{5, 3\}), one max pooling layer (kernel size 2, stride 2), and two fully connected layers to produce a 32-dimensional depth feature. This feature is concatenated with the measured proprioception and passed into a GRU with a 512-unit hidden size to infer the privileged information from historical data.

\subsection*{S7. Human-robot collaboration}\label{secS7}

\begin{figure}
	\centering
	\includegraphics[width=0.7\textwidth]{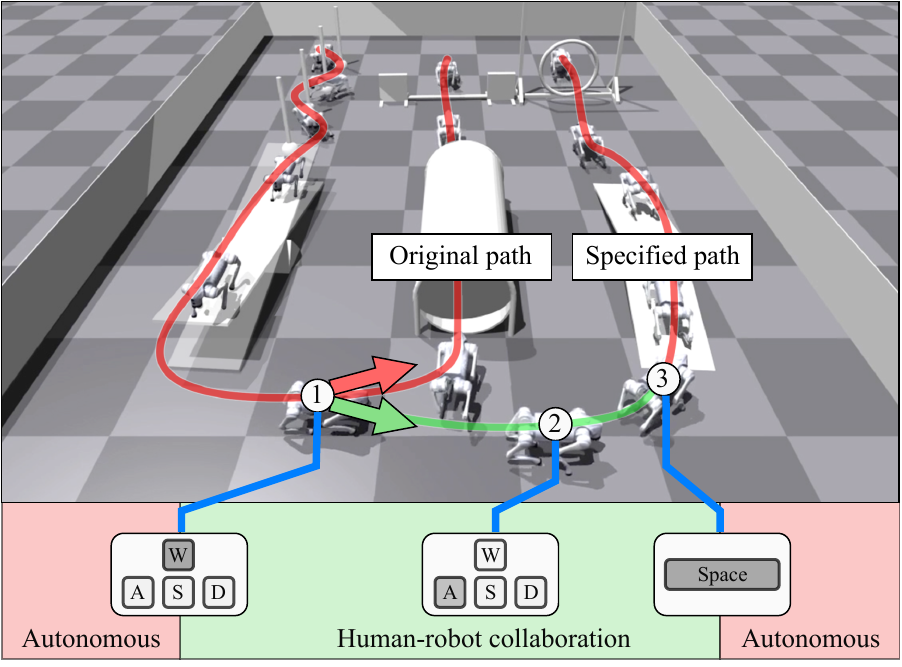}
	\caption{\textbf{Human-robot collaboration.} The robot receives the operator’s input at point 1, where it switches from an immediate left turn to moving straight. The operator only adjusts the linear and angular velocities, while the robot's behavior style remains controlled by the TSC. At point 3, the operator presses the Space key, and the robot transitions from human-robot collaboration to autonomous movement.}
	\label{fig:human_robot}
\end{figure}

In the quadrupedal agility challenge, the robot follows a specific route through randomly placed obstacles. However, in situations where the route is ambiguous, human collaboration is needed to complete the task, similar to the role of handlers in dog agility, where handlers are informed of the route in advance. We simulate this collaborative mode in a scenario with bidirectional routes (Fig.~\ref{fig:human_robot}). The robot receives the operator’s input at point 1, where it switches from an immediate left turn to moving straight. The operator only adjusts the linear and angular velocities, while the robot's behavior style remains controlled by the TSC. At point 3, the operator presses the Space key, and the robot transitions from human-robot collaboration to autonomous movement. We use a remote control for this task, though other methods could also achieve the same goal, such as voice commands, gestures, or behavioral intentions, which we plan to explore in future work. In addition to altering the route, the handler can also modify the skill variable inputs to switch between different behaviors.

\subsection*{S8. Ablation of the BYOL loss for student training}\label{secS8}

\begin{figure}[ht]
	\centering
	\includegraphics[width=0.7\textwidth]{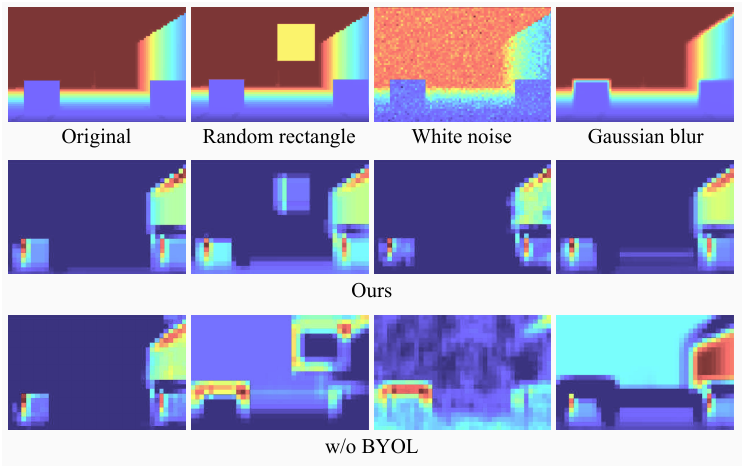}
	\caption{\textbf{Visualization of the depth image encoder's focused areas.} The first row shows the original depth image, followed by depth images with added random rectangles, white noise, and Gaussian blur. The second row displays the areas of focus for our method's encoder. The third row shows the areas of focus for the encoder without BYOL.}
	\label{fig:tsc_heat_map}
\end{figure}

To intuitively demonstrate the improvement in the robustness of the student policy by the BYOL loss, we use Gradient-weighted Class Activation Mapping (Grad-CAM)~\cite{selvaraju2020grad} to visualize the areas of focus in the depth image encoder, as shown in Fig.~\ref{fig:tsc_heat_map}. The first row shows the original depth image containing the bar-jump, as well as the depth images with added random rectangles, white noise, and Gaussian blur. The second row displays the areas of focus for our method's encoder. The third row shows the areas of focus for the encoder without BYOL. With the use of BYOL, the encoder consistently focuses on important areas under various noise disturbances, ensuring the policy's robustness in real-world scenarios.

\begin{figure}[ht]
	\centering
	\includegraphics[width=0.8\textwidth]{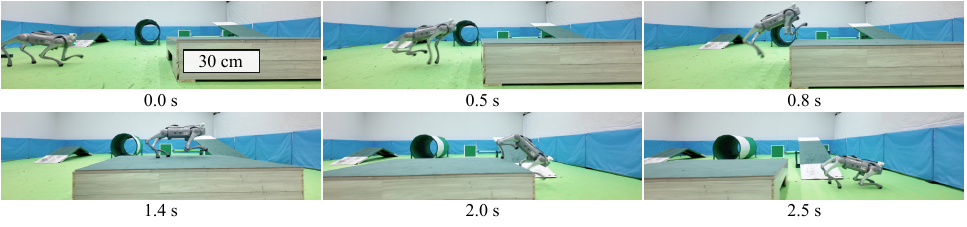}
	\caption{\textbf{Box jumping process.} The box is 30 cm high, approximately the same as the robot's normal standing height. With our controller, the quadrupedal robot can quickly jump onto the obstacle without relying on its front legs for support.}
	\label{fig:box_real}
\end{figure}

\begin{figure}[ht]
	\centering
	\includegraphics[width=0.8\textwidth]{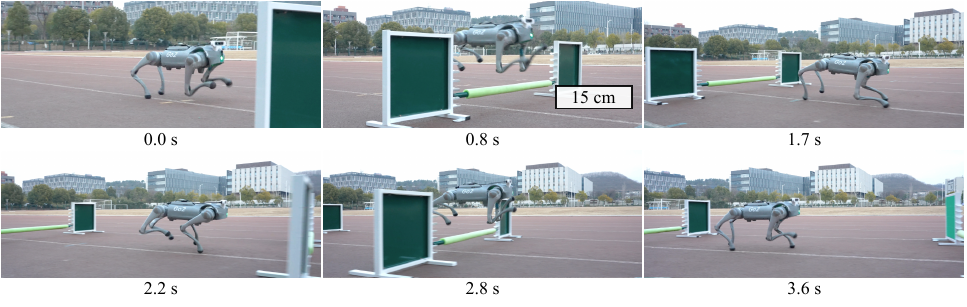}
	\caption{\textbf{Hurdling process.} We showcase two jumping phases in the hurdling task. The robot utilizes depth vision to naturally switch behaviors, completing the task like a real dog. The whole process can be found in Supplementary Video 3.}
	\label{fig:hurdle_real}
\end{figure}

\begin{table}
	\centering
	\caption{\textbf{Hyperparameters for semi-supervised InfoGAIL.}}
	\label{tab:hype_ss_infogail}
	\begin{tabular}{ll}
		\\
		\hline
		Optimizer & Adam\\
		Policy \& value learning rate & 1.0 E-3\\
        Discriminator learning rate & 5.0 E-4\\
		Encoder learning rate & 1.0 E-3\\
        Observation horizon & 2\\
        Discount factor & 0.99\\
        TD($\lambda$) & 0.95\\
        PPO clip threshold & 0.2\\
        Moving average weight & 1.0 E-3\\
        Imitation reward weight & 0.2\\
        Task reward weight & 0.2\\
		\hline
	\end{tabular}
\end{table}

\begin{table}
	\centering
	\caption{\textbf{Task command ranges for quadrupedal agility challenge.}}
	\label{tab:command_range}
	\begin{tabular}{cccccc}
		\\
		\hline
		 & Walk & Pace & Trot & Canter & Jump\\
        \hline
		$v_x$ (m/s) & [0.0, 0.6] & [0.5, 1.5] & [0.5, 1.5] & [0.8, 2.0] & [0.8, 1.8]\\
		$v_y$ (m/s) & [-0.15, 0.15] & [-0.3, 0.3] & [-0.3, 0.3] & [-0.5, 0.5] & [-0.3, 0.3]\\
		$\omega_z$ (rad/s) & [-1.0, 1.0] & [-1.57, 1.57] & [-1.57, 1.57] & [-0.5, 0.5] & [-0.5, 0.5]\\
		jump height (m) & - & - & - & - & [0.45, 0.55]\\
        locomotion height (m) & [0.25, 0.34] & [0.25, 0.34] & [0.25, 0.34] & [0.25, 0.34] & -\\
		\hline
	\end{tabular}
\end{table}

\begin{table}
	\centering
	\caption{\textbf{Hyperparameters for EASI.}}
	\label{tab:hype_easi}
	\begin{tabular}{ll}
		\\
		\hline
		Optimizer & RMSprop\\
		Discriminator learning rate & 3.0 E-4\\
            Discriminator training epoch  & 10 \\
            Number of env. & 300\\
            Evolution survival rate & 0.5\\
            Trajectory length & 500\\

		\hline
	\end{tabular}
\end{table}

\begin{table}
	\centering
	\caption{\textbf{Hyperparameters for privileged learning.}}
	\label{tab:hype_priv}
	\begin{tabular}{ll}
		\\
		\hline
		Optimizer & Adam\\
		Policy \& value learning rate & 5.0 E-4\\
        Discount factor & 0.99\\
        TD($\lambda$) & 0.95\\
        PPO clip threshold & 0.2\\
        Imitation reward weight & 0.2\\
        Task reward weight & 2.0\\
		\hline
	\end{tabular}
\end{table}

\begin{algorithm}[H]
\caption{BBC training with semi-supervised InfoGAIL}
\label{alg:ss_infogail}
\begin{algorithmic}[1]
    \State Initialize policy $\pi_{0}$, value function $V_{0}$, discriminator $D_0$, predictor $Q_{0}$, and latent skill distribution $p_{0}(\mathbf{c})$. Prepare unlabeled and labeled demonstrations $d^{\text{E}}$ and $d^{\text{EL}}$
    \For{$i=0, 1, 2, \cdots N$}
        \State Sample a batch of latent variables $\mathbf{c} \sim p_{i}(\mathbf{c})$, $\epsilon \sim p(\epsilon)$
        \State Interact with the environment using $\pi_{i}$, $\mathbf{c}$ and $\epsilon$, and obtain $d^{\pi}$
        \State Sample a batch of observations $b^{\text{E}} \sim d^{\text{E}}$, $b^{\text{EL}} \sim d^{\text{EL}}$, $b^{\pi} \sim d^{\pi}$
        \State Update $D_{i}$ to $D_{i+1}$ and $Q_{i}$ to $Q_{i+1}$ according to Equation~(\ref{eq_ss_infogail})
        \State Update $\pi_{i}$ to $\pi_{i+1}$ using PPO~\cite{schulman2017proximal} with reward according to Equation~(\ref{eq_reward})
        \State Update $V_{i}$ to $V_{i+1}$ using $TD(\lambda)$
        \State Update $p_{i}(\mathbf{c})$ to $p_{i+1}(\mathbf{c})$ with $Q_{i+1}$ and $b^{\text{E}}$
    \EndFor
\end{algorithmic}
\end{algorithm}

\begin{algorithm}[H]
\caption{Simulator parameter optimization with EASI}
\label{alg:easi}
\begin{algorithmic}[1]
    \State Initialize parameter distribution $\Xi^{(0)}$, discriminator $\tilde{D}_0$, policy $\pi$, real-world demonstration $\tilde{d}^{E}$
    \For{generation $i=0, 1, 2, \cdots G$}
        \State Sample individuals by $\xi^{(i)}_{j}\sim\Xi^{(i)}$, $j=1, 2, \cdots, N$
        \For{simulation environment $j=1, 2, \cdots, N$}
            \State Set simulator parameters to $\xi^{(i)}_{j}$
            \State Use $\pi$ to sample trajectory $\tau_j$
            \State Store $\tau_j$ in $d^{\pi}$
        \EndFor
        \For{update step $k=1, 2, \cdots, K$}
            \State Sample a batch of transitions $\tilde{b}^{E}\sim\tilde{d}^{E}$, $b^{\pi} \sim d^{\pi}$
            \State Update $\tilde{D}_{k}$ to $\tilde{D}_{k+1}$ according to Equation~(\ref{eq_easi})
        \EndFor
        \State Calculate discriminator reward $r^{(i)}$ according to Equation~(\ref{eq_r_gail})
        \State Using ES to find next generation distribution $\Xi^{(i+1)}=\text{ES}(\xi^{(i)}, r^{(i)})$
    \EndFor
\end{algorithmic}
\end{algorithm}

\begin{algorithm}[H]
\caption{TSC training with privileged learning}
\label{alg:priv}
\begin{algorithmic}[1]
    \State Initialize teacher policy $\pi^{\text{TSC}}_{0}$, student policy $\hat{\pi}^{\text{TSC}}_{0}$, value function $V_{0}$
    \For{$i=0, 1, 2, \cdots N$}
        \State Initialize obstacles and robot at a random position
        \State Interact with the environment using $\pi^{\text{TSC}}_{i}$, and obtain $d^{\pi}$
        \State Update $\pi^{\text{TSC}}_{i}$ to $\pi^{\text{TSC}}_{i+1}$ using Hybrid-PPO~\cite{ijcai2019p0316} with reward according to Equation~(\ref{eq_reward_TSC})
        \State Update $V_{i}$ to $V_{i+1}$ using $TD(\lambda)$
    \EndFor
    \For{$j=0, 1, 2, \cdots M$}
        \State Initialize obstacles and robot at a random position
        \State Interact with the environment using $\hat{\pi}^{\text{TSC}}_{j}$, and obtain $\hat{d}^{\pi}$
        \State Update $\hat{\pi}^{\text{TSC}}_{j}$ to $\hat{\pi}^{\text{TSC}}_{j+1}$ by descending with gradients according to Equation~(\ref{eq_tsc_student})
    \EndFor
\end{algorithmic}
\end{algorithm}

\end{document}